# Literal Movement Grammars

**Annius V. Groenink**[*]
CWI
Kruislaan 413
1098 SJ Amsterdam
The Netherlands
avg@cwi.nl

## Abstract

*Literal movement grammars* (**LMG**s) provide a general account of extraposition phenomena through an attribute mechanism allowing top-down displacement of syntactical information. **LMG**s provide a simple and efficient treatment of complex linguistic phenomena such as cross-serial dependencies in German and Dutch—separating the treatment of natural language into a parsing phase closely resembling traditional context-free treatment, and a disambiguation phase which can be carried out using matching, as opposed to full unification employed in most current grammar formalisms of linguistical relevance.

## 1 Introduction

The motivation for the introduction of the *literal movement grammars* presented in this paper is twofold. The first motivation is to examine whether, and in which ways, the use of unification is essential to automated treatment of natural language. Unification is an expensive operation, and pinpointing its precise role in **NLP** may give access to more efficient treatment of language than in most (Prolog-based) scientific applications known today. The second motivation is the desire to apply popular computer-science paradigms, such as the theory of attribute grammars and modular equational specification, to problems in linguistics. These formal specification techniques, far exceeding the popular *Prolog* in declarativity, may give new insight into the formal properties of natural language, and facilitate prototyping for large language applications in the same way as they are currently being used to facilitate prototyping of programming language tools.

For an extensive illustration of how formal specification techniques can be made useful in the treatment of natural language, see (Newton, 1993) which describes the abstract specification of several accounts of phrase structure, features, movement, modularity and parametrization so as to abstract away from the exact language being modelled. The specification language (ASL) used by Newton is a very powerful formalism. The class of specification formalisms we have in mind includes less complex, equational techniques such as ASF+SDF (Bergstra et al., 1989) (van Deursen, 1992) which can be applied in practice by very efficient execution as a term rewriting system.

Literal movement grammars are a straightforward extension of context-free grammars. The derivation trees of an **LMG** analysis can be easily transformed into trees belonging to a context-free backbone which gives way to treatment by formal specification systems. In order to obtain an efficient implementation, some restrictions on the general form of the formalism are necessary.

### 1.1 Structural Context Sensitivity in Natural Language

Equational specification systems such as the ASF+SDF system operate through sets of *equations* over signatures that correspond to arbitrary forms of context-free grammar. An attempt at an equational specification of a grammar based on context-free phrase structure rules augmented with feature constraints may be to use the context-free backbone as a signature, and then implement further analysis through equations over this signature. This seems entirely analogous to the static semantics of a programming language: the language itself is context-free, and the static semantics are defined in terms of functions over the constructs of the language.

In computer-science applications it is irrelevant whether the evaluation of these functions is carried out during the parsing phase (*1-pass treatment*), or afterwards (*2-pass treatment*). This is not a trivial property of computer languages: a computer language with static semantics restrictions is a context-sensitive sublanguage of a context-free language that is either *unambiguous* or has the *finite ambiguity* property: for any input sentence, there is only a finite number of possible context-free analyses.

In section 1.3 we will show that due to phenomena of *extraposition* or *discontinuous constituency* exhibited by natural languages, a context-free backbone for a sufficiently rich fragment of natural language no longer has the property of finite ambiguity. Hence an initial stage of sentence processing cannot be based on a purely context-free analysis.

[*]This work is supported by SION grant 612-317-420 of the Netherlands Organization for Scientific Research (NWO).

The **LMG** formalism presented in this paper attempts to eliminate infinite ambiguity by providing an elementary, but adequate treatment of movement. Experience in practice suggests that after relocating displaced constituents, a further analysis based on feature unification no longer exploits unbounded structural embedding. Therefore it seems that after **LMG**-analysis, there is no need for unification, and further analysis can be carried out through functional matching techniques.

### 1.2 Aims

We aim to present a grammar formalism that

▷ is sufficiently powerful to model relevant fragments of natural language, at least large enough for simple applications such as an interface to a database system over a limited domain.

▷ is sufficiently elementary to act as a front-end to computer-scientific tools that operate on context-free languages.

▷ has a (sufficiently large) subclass that allows efficient implementation through standard (Earley-based) left-to-right parsing techniques.

### 1.3 Requirements

Three forms of movement in Dutch will be a leading thread throughout this paper. We will measure the adequacy of a grammar formalism in terms of its ability to give a unified account of these three phenomena.

**Topicalization** The (leftward) movement of the objects of the verb phrase, as in

(1)  [Which book]$_i$  did John forget to return $e_i$ to the library?

**Dutch sentence structure** The surface order of sentences in Dutch takes three different forms: the finite verb appears inside the verb phrase in relative clauses; before the verb phrase in declarative clauses, and before the subject in questions:

(2)  ...dat Jan [$_{VP}$ Marie kuste ]

(3)  Jan kuste$_i$ [$_{VP}$ Marie $e_i$ ]

(4)  kuste$_i$ Jan [$_{VP}$ Marie $e_i$ ]?

We think of these three (surface) forms as merely being different representations of the same (deep) structure, and will take this deep structure to be the form (2) that does not show movement.

**Cross-serial dependencies** In Dutch and German, it is possible to construct sentences containing arbitrary numbers of crossed dependencies, such as in

(5)  ... dat  Marie Jan$_i$ Fred$_j$ Anne$_k$
that
hoorde$_i$   helpen$_j$   overtuigen$_k$
heard        help         convince

(that Mary heard John help Fred convince Anne). Here the $i, j, k$ denote which noun is the first object of which verb. The analysis we have in mind for this is

dat Marie Jan$_i$ Fred$_j$ Anne$_k$
[$_{VP}$ hoorde $e_i$  helpen $e_j$  overtuigen $e_k$.]

Note that this analysis (after relocation of the extraposed objects) is structurally equal to the corresponding English *VP*. The accounts of Dutch in this paper will consistently assign "deep structures" to sentences of Dutch which correspond to the underlying structure as it appears in English. Similar accounts can be given for other languages—so as to get a uniform treatment of a group of similar (European) languages such as German, French and Italian.

If we combine the above three analyses, the final analysis of (3) will become

Jan kuste$_i$ Marie$_j$ [$_{VP}$ $e_i$ $e_j$ ]

Although this may look like an overcomplication, this abundant use of movement is essential in any uniform treatment of Dutch verb constructions. Hence it turns out to occur in practice that a verb phrase has no lexical expansion at all, when a sentence shows both object and verb extraposition. Therefore, as conjectured in the introduction, a 2-pass treatment of natural language based on a context-free backbone will in general fail—as there are infinitely many ways of building an empty verb phrase from a number of empty constituents.

## 2 Definition and Examples

There is evidence that suggests that the typical human processing of movement is to *first* locate displaced information (the *filler*), and *then* find the logical location (the *trace*), to substitute that information. It also seems that by and large, displaced information appears earlier than (or left of) its logical position, as in all examples given in the previous section. The typical unification-based approach to such movement is to structurally analyse the displaced constituent, and use this analysed information in the treatment of the rest of the sentence. This method is called *gap-threading*; see (Alshawi, 1992).

If we bear in mind that a filler is usually found to the left of the corresponding trace, it is worth taking into consideration to develop a way of *deferring treatment of syntactical data*. E.g. for example sentence 1 this means that upon finding the displaced constituent *which book*, we will not evaluate that constituent, but rather remember during the treatment of the remaining part of the sentence, that this data is still to be fitted into a logical place.

This is not a new idea. A number of *non-concatenative* grammar formalisms has been put forward, such as *head-wrapping grammars* (**HG**) (Pollard, 1984), *extraposition grammars* (**XG**) (Pereira, 1981). and *tree adjoining grammars* (**TAG**) (Kroch and Joshi, 1986). A discussion of these formalisms as alternatives to the **LMG** formalism is given in section 4.

Lessons in parsing by hand in high school (e.g. in English or Latin classes) informally illustrate the purpose of *literal movement grammars*: as opposed to the traditional linguistic point of view that there is only one *head* which dominates a phrase, constituents of a sentence have *several key components*. A verb phrase for example not only has its finite verb, but also one or more objects. It is precisely these key components that can be subject to movement. Now when such a key component is found outside the consitituent it belongs to, the **LMG** formalism implements a simple mechanism to pass the component down the derivation tree, where it is picked up by the constituent that contains its trace.

It is best to think of **LMG**s versus context-free grammars as a predicate version of the (propositional) paradigm of context-free grammars, in that nonterminals can have arguments. If we call the general class of such grammars *predicate grammars*, the distinguishing feature of **LMG** with respect to other predicate grammar formalisms such as *indexed grammars*[1] (Weir, 1988) (Aho, 1968) is the ability of *binding* or *quantification* in the right hand side of a phrase structure rule.

▎**2.1 Definition** We fix disjoint sets $\mathbf{N}, \mathbf{T}, \mathbf{V}$ of *nonterminal symbols*, *terminal symbols* and *variables*. We will write $A, B, C \ldots$ to denote nonterminal symbols, $a, b, c \ldots$ to denote terminal symbols, and $x, y, z$ for variables. A sequence $a_1 a_2 \cdots a_n$ or $\boldsymbol{a} \in \mathbf{T}^*$ is called a (terminal) *word* or *string*. We will use the symbols $\boldsymbol{a}, \boldsymbol{b}, \boldsymbol{c}$ for terminal words. (Note the use of bold face for sequences.)

▎**2.2 Definition (term)** A sequence $t_1 t_2 \cdots t_n$ or $\boldsymbol{t} \in (\mathbf{V} \cup \mathbf{T})^*$ is called a *term*. If a term consists of variables only, we call it a *vector* and usually write $\boldsymbol{x}$.

▎**2.3 Definition (similarity type)** A (partial) function $\mu$ mapping $\mathbf{N}$ to the natural numbers is called a *similarity type*.

▎**2.4 Definition (predicate)** Let $\mu$ be a similarity type, $A \in \mathbf{N}$ and $n = \mu(A)$, and for $1 \leq i \leq n$, let $t_i$ be a term. Then a *predicate* $\varphi$ of type $\mu$ is a terminal $a$ (a *terminal predicate*) or a syntactical unit of the form $A(t_1, t_2, \ldots, t_n)$, called a *nonterminal predicate*. If all $\boldsymbol{t}_i = \boldsymbol{x}_i$ are vectors, we say that $\varphi = A(\boldsymbol{x}_1, \boldsymbol{x}_2, \ldots, \boldsymbol{x}_n)$ is a *pattern*.

Informally, we think of the arguments of a nonterminal as terminal words. A predicate $A(x)$ then stands for a constituent $A$ where certain information with terminal yield $x$ has been extraposed (i.e. found outside the constituent), and must hence be left out of the $A$ constituent itself.

▎**2.5 Definition (item)** Let $\mu$ be a similarity type, $\varphi$ a predicate of type $\mu$, and $\boldsymbol{t}$ a term. Then an *item* of type $\mu$ is a syntactical unit of one of the following forms:

1. $\varphi$ (a nonterminal or terminal predicate)
2. $x{:}\varphi$ (a *quantifier* item)
3. $\varphi/\boldsymbol{t}$ (a *slash* item)

We will use $\Phi, \Psi$ to denote items, and $\alpha, \beta, \gamma$ to denote sequences of items.

▎**2.6 Definition** Let $\mu$ be a similarity type. A *rewrite rule* $R$ of type $\mu$ is a syntactical unit $\varphi \to \Phi_1 \Phi_2 \cdots \Phi_n$ where $\varphi$ is a pattern of type $\mu$, and for $1 \leq i \leq n$, $\Phi_i$ is an item of type $\mu$.

A *literal movement grammar* is a triple $(\mu, S, P)$ where $\mu$ is a similarity type, $S \in \mathbf{N}$, $\mu(S) = 0$ and $P$ is a set of rewrite rules of type $\mu$.

Items on the right hand side of a rule can either *refer to* variables, as in the following rule:

$$A(x, yz) \to B()/x \;\; \mathtt{a}/y \;\; C(z)$$

or *bind* new variables, as the first two items in

$$A() \to x{:}B() \;\; y{:}C(x) \;\; D(y).$$

A slash item such as $B()/x$ means that $x$ should be used instead of the actual "input" to recognize the nonterminal predicate $B()$. I.e. the terminal word $x$ should be recognized as $B()$, and the item $B()/x$ itself will recognize the empty string. A *quantifier* item $x{:}B()$ means that a constituent $B()$ is recognized from the input, and the variable $x$, when used elsewhere in the rule, will stand for the part of the input recognized.

▎**2.7 Definition (rewrite semantics)** Let $R = A(\boldsymbol{x}_1, \ldots, \boldsymbol{x}_n) \to \Phi_1 \Phi_2 \cdots \Phi_m$ be a rewrite rule, then an *instantiation* of $R$ is the syntactical entity obtained by substituting for each $i$ and for each variable $x \in \boldsymbol{x}_i$ a terminal word $\boldsymbol{a}_x$.

A grammar derives the string $\boldsymbol{a}$ iff $S() \stackrel{G}{\Longrightarrow} \boldsymbol{a}$ where $\stackrel{G}{\Longrightarrow}$ is a relation between predicates and sequences of items defined inductively by the following axioms and inference rules:[2]

$$\boldsymbol{a} \stackrel{G}{\Longrightarrow} \boldsymbol{a}$$

$$\varphi \stackrel{G}{\Longrightarrow} \alpha \quad \text{when } \varphi \to \alpha \text{ is an instantiation of a rule in } G$$

$$\frac{\varphi \stackrel{G}{\Longrightarrow} \beta \; A(\boldsymbol{t}_1, \ldots, \boldsymbol{t}_n) \; \gamma \quad A(\boldsymbol{t}_1, \ldots, \boldsymbol{t}_n) \stackrel{G}{\Longrightarrow} \boldsymbol{a}}{\varphi \stackrel{G}{\Longrightarrow} \beta \; \boldsymbol{a} \; \gamma} \textbf{MP}$$

$$\frac{\varphi \stackrel{G}{\Longrightarrow} \beta \; \psi/\boldsymbol{a} \; \gamma \quad \psi \stackrel{G}{\Longrightarrow} \boldsymbol{a}}{\varphi \stackrel{G}{\Longrightarrow} \beta \; \gamma} \textbf{/E}$$

$$\frac{\varphi \stackrel{G}{\Longrightarrow} \beta \; x{:}\psi \; \gamma \quad \psi \stackrel{G}{\Longrightarrow} \boldsymbol{a}}{\varphi \stackrel{G}{\Longrightarrow} (\beta \; \boldsymbol{a} \; \gamma)[\boldsymbol{a}/x]} \textbf{:E}$$

---

[1] Indexed grammars are a weak form of *monadic* predicate grammar, as a nonterminal can have at most one argument.

[2] Note that $[\boldsymbol{a}/x]$ in the **:E** rule is not an item, but stands for the substitution of $\boldsymbol{a}$ for $x$.

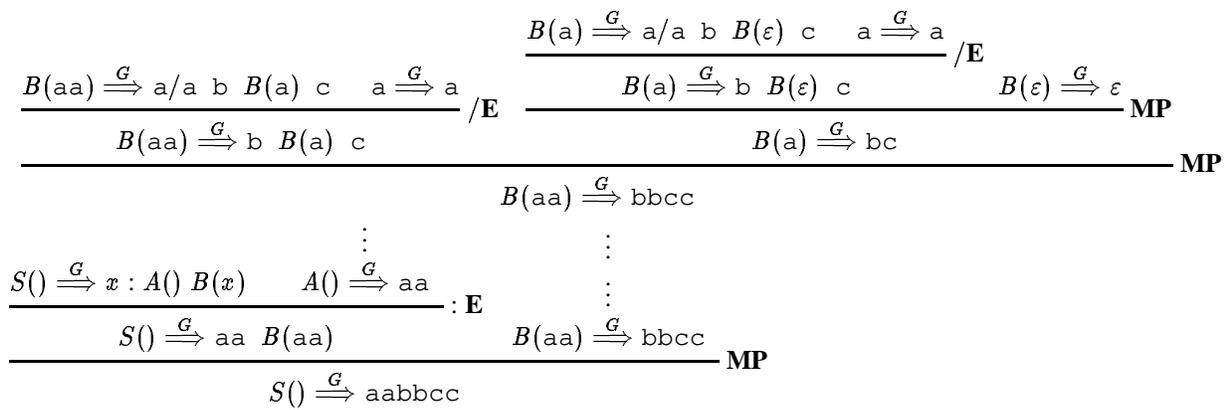

Figure 1. Derivation of `aabbcc`.

▌**2.8 Example ($a^n b^n c^n$)** The following, very elementary **LMG** recognizes the trans-context free language $a^n b^n c^n$:

$$
\begin{array}{lcl}
S() & \rightarrow & x{:}A()\ B(x) \\
A() & \rightarrow & a\ A() \\
A() & \rightarrow & \varepsilon \\
B(xy) & \rightarrow & a/x\ b\ B(y)\ c \\
B(\varepsilon) & \rightarrow & \varepsilon
\end{array}
$$

Figure 1 shows how `aabbcc` is derived according to the grammar. The informal tree analysis in figure 2

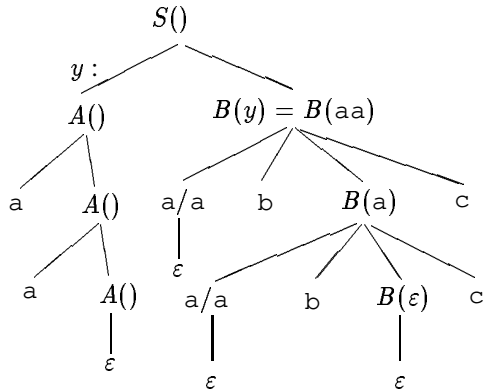

Figure 2. Informal tree analysis.

illustrates more intuitively how displaced information (the two `a` symbols in this case) is 'moved back down' into the tree, until it gets 'consumed' by a slash item. It also shows how we can extract a context-free 'deep structure' for further analysis by, for example, formal specification tools: if we transform the tree, as shown in figure 3, by removing quantified (extraposed) data, and abstracting away from the parameters, we see that the grammar, in a sense, works by transforming the language $a^n b^n c^n$ to the context-free language $(ab)^n c^n$. Figure 4 shows how we can derive a context free 'backbone grammar' from the original grammar.

▌**2.9 Example (cross-serial dependencies in Dutch)** The following **LMG** captures precisely the three basic types of extraposition defined in section 1.3: the three Dutch verb orders, topicalization and cross-serial verb-object dependencies.

$$
\begin{array}{lcl}
S & \rightarrow & S'(\varepsilon) \\
S'(\varepsilon) & \rightarrow & \text{dat}\ NP\ VP(\varepsilon, \varepsilon) \\
S'(\varepsilon) & \rightarrow & n{:}NP\ S'(n) \\
S'(n) & \rightarrow & v{:}V\ NP\ VP(v, n) \\
S'(\varepsilon) & \rightarrow & NP\ v{:}V\ VP(v, \varepsilon) \\
VP(v, n) & \rightarrow & m{:}NP\ VP(v, nm) \\
VP(v, n) & \rightarrow & V'(v, n) \\
V'(\varepsilon, \varepsilon) & \rightarrow & VI \\
V'(v, \varepsilon) & \rightarrow & VI/v \\
V'(\varepsilon, n) & \rightarrow & VT\ NP/n \\
V'(v, n) & \rightarrow & VT/v\ NP/n \\
V'(\varepsilon, nm) & \rightarrow & VR\ NP/n\ V'(\varepsilon, m) \\
V'(v, nm) & \rightarrow & VR/v\ NP/n\ V'(\varepsilon, m) \\
V & \rightarrow & VI \\
V & \rightarrow & VT \\
V & \rightarrow & VR
\end{array}
$$

A sentence $S'$ has one argument which is used, if nonempty, to fill a noun phrase trace. A *VP* has two

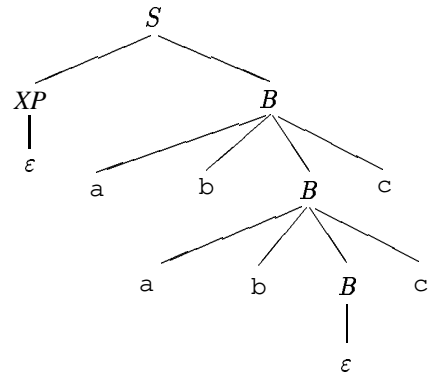

Figure 3. Context free backbone.

$$
\begin{aligned}
S &\rightarrow XP\ B \\
B &\rightarrow a\ b\ B\ c \\
B &\rightarrow \varepsilon \\
XP &\rightarrow \varepsilon
\end{aligned}
$$

Figure 4. Backbone grammar.

arguments: the first is used to fill verb traces, the second is treated as a list of noun phrases to which more noun phrases can be appended. A *V'* is similar to a *VP* except that it uses the list of noun phrases in its second argument to fill noun phrase traces rather than adding to it.

Figure 5 shows how this grammar accepts the sentence

*Marie zag Fred Anne kussen.*

We see that it is analyzed as

*Marie zag$_i$ Fred$_j$ Anne$_k$*
    $[_{V'}\ e_i\ e_j\ [_{V'}\ kussen\ e_k\ ]]$

which as anticipated in section 1.3 has precisely the basic, context-free underlying structure of the corresponding English sentence *Mary saw Fred kiss Anne* indicated in figure 5 by terminal words in bold face. Note that arbitrary verbs are recognized by a quanti-fier item $v{:}V$, and only when, further down the tree, a trace is filled with such a verb in items such as $VR/v$, its subcategorization types *VI*, *VT* and *VR* start playing a role.

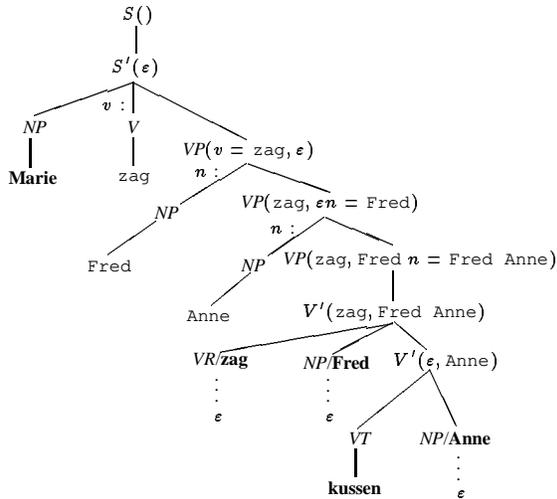

Figure 5. Derivation of a Dutch sentence

## 3 Formal Properties

The **LMG** formalism in its unrestricted form is shown to be Turing complete in (Groenink, 1995a). But the grammars presented in this paper satisfy a number of vital properties that allow for efficient parsing techniques.

Before building up material for a complexity result, notice the following proposition, which shows, using only part of the strength of the formalism, that the literal movement grammars are closed under intersection.

**▌3.1 Proposition (intersection)** Given two literal movement grammars $G_1 = (\mu_1, S_1, P_1)$ and $G_2 = (\mu_2, S_2, P_2)$ such that $\text{dom}(\mu_1) \cap \text{dom}(\mu_2) = \emptyset$, we can construct the grammar $G_I = (\mu_1 \cup \mu_2 \cup \{(S,0)\}, S, P_1 \cup P_2 \cup \{R\})$ where we add the rule $R$:

$$S() \rightarrow x{:}S_1()\ S_2()/x$$

Clearly, $G_I$ recognizes precisely those sentences which are recognized by both $G_1$ and $G_2$.

We can use this knowledge in example 2.9 to restrict movement of verbs to verbs of finite morphology, by adding a nonterminal *VFIN*, replacing the quantifier items $v{:}V$ that locate verb fillers with $v{:}VFIN$, where *VFIN* generates all finite verbs. Any extraposed verb will then be required to be in the intersection of *VFIN* and one of the verb types *VI*, *VT* or *VR*, reducing possible ambiguity and improving the efficiency of left-to-right recognition.

The following properties allow us to define restrictions of the **LMG** formalism whose recognition problem has a polynomial time complexity.

**▌3.2 Definition (non-combinatorial)** An **LMG** is *non-combinatorial* if every argument of a nonterminal on the RHS of a rule is a single variable (i.e. we do not allow composite terms within predicates). If $G$ is a non-combinatorial **LMG**, then any terminal string occurring (either as a sequence of items or inside a predicate) in a full $G$-derivation is a substring of the derived string. The grammar of example 2.8 is non-combinatorial; the grammar of example 2.9 is not (the offending rule is the first *VP* production).

**▌3.3 Definition (left-binding)** An **LMG** $G$ is *left-binding* when

1. W.r.t. argument positions, an item in the RHS of a rule only depends on variables bound in items to its left.
2. For any vector $x_1 \cdots x_n$ of $n > 1$ variables on the LHS, each of $x_1$ upto $x_{n-1}$ occurs in exactly one item, which is of the form $\varphi/x_i$. Furthermore, for each $1 \leq l < k \leq n$ the item referring to $x_l$ appears left of any item referring to $x_k$.

For example, the following rule is left binding:

$$A(xyz, v) \rightarrow u{:}B(v)\ C(v)/x\ D()/y\ E(u,z)$$

but these ones are not:

(a) $A(y) \rightarrow C(x)\ x{:}D(y)$
(b) $A(xy) \rightarrow A(x)\ B(y)$
(c) $A(xyz) \rightarrow A(z)\ B()/x\ C()/y$

because in (**a**), $x$ is bound right of its use; in (**b**), the item $A(x)$ is not of the form $\varphi/x$ and in (**c**), the variables in the vector $xyz$ occur in the wrong order ($zxy$).

If a grammar satisfies condition 1, then for any derivable string, there is a derivation such that the modus ponens and elimination rules are always applied to the leftmost item that is not a terminal. Furthermore, the **:E** rule can be simplified to

$$\frac{\varphi \overset{G}{\Rightarrow} \beta \ x{:}\psi \ \gamma \quad \psi \overset{G}{\Rightarrow} \boldsymbol{a}}{\varphi \overset{G}{\Rightarrow} \beta \ \boldsymbol{a} \ (\gamma[\boldsymbol{a}/x])} : \mathbf{E}$$

The proof tree in example 2.8 (figure 1) is an example of such a derivation.

Condition 2 eliminates the nondeterminism in finding the right instantiation for rules with multiple variable patterns in their LHS.

Both grammars from section 2 are left-binding.

■ **3.4 Definition (left-recursive)** An **LMG** $G$ is *left-recursive* if there exists an instantiated nonterminal predicate $\varphi$ such that there is a derivation of $\varphi \overset{G}{\Rightarrow} \varphi\alpha$ for any sequence of items $\alpha$.

The following two rules show that left-recursion in **LMG** is not always immediately apparent:

$$\begin{array}{rcl} A(y) & \to & B()/y \ A(\varepsilon) \\ B() & \to & \varepsilon \end{array}$$

for we have

$$\frac{A(\varepsilon) \overset{G}{\Rightarrow} B()/\varepsilon \ A(\varepsilon) \quad B() \overset{G}{\Rightarrow} \varepsilon}{A(\varepsilon) \overset{G}{\Rightarrow} A(\varepsilon)} / \mathbf{E}$$

We now show that the recognition problem for an arbitrary left-binding, non-combinatorial **LMG** has a polynomial worst-case time complexity.

■ **3.5 Theorem (polynomial complexity)** Let $G$ be a **LMG** of similarity type $\mu$ that is non-combinatorial, left binding and not left-recursive. Let $m$ be the maximum number of items on the right hand side of rules in $G$, and let $p$ be the greatest arity of predicates occurring in $G$. Then the worst case time complexity of the recognition problem for $G$ does not exceed $\mathcal{O}(|G|m(1+p)n^{1+m+2p})$, where $n$ is the size of the input string $a_1 a_2 \cdots a_n$.

**Proof** (sketch) We adopt the memoizing recursive descent algorithm presented in (Leermakers, 1993). As $G$ is not left-binding, the terminal words associated with variables occurring in the grammar rules can be fully determined while proceeding through sentence and rules from left to right. Because the grammar is non-combinatorial, the terminal words substituted in the argument positions of a nonterminal are always substrings of the input sentence, and can hence be represented as a pair of integers.

The recursive descent algorithm recursively computes set-valued recognition functions of the form:

$$[\varphi](i) = \{ j \mid \varphi \overset{G}{\Rightarrow} a_{i+1} \cdots a_j \}$$

where instead of a nonterminal as in the context-free case, $\varphi$ is any instantiated nonterminal predicate $A(\boldsymbol{b}_1, \ldots, \boldsymbol{b}_n)$. As $\boldsymbol{b}_1, \ldots, \boldsymbol{b}_n$ are continuous substrings of the input sentence $a_1 a_2 \cdots a_n$, we can reformulate this as

$$\begin{aligned} & [A](i, (l_1, r_1), \ldots, (l_\mu, r_\mu)) \\ & = \ \{ j \mid A(a_{l_1+1} \cdots a_{r_1}, \ldots, a_{l_\mu+1} \cdots a_{r_\mu}) \\ & \qquad \overset{G}{\Rightarrow} a_{i+1} \cdots a_j \} \end{aligned}$$

Where $\mu = \mu(A) \leq p$. The arguments $i, l_1, \ldots, l_\mu$ and $r_1, \ldots, r_\mu$ are integer numbers ranging from 0 to $n-1$ and 1 to $n$ respectively. Once a result of such a recognition function has been computed, it is stored in a place where it can be retrieved in one atomic operation. The number of such results to be stored is $\mathcal{O}(n)$ for each possible nonterminal and each possible combination of, at most $1 + 2p$, arguments; so the total space complexity is $\mathcal{O}(|\mathbf{G}|n^{2+2p})$.

Much of the extra complication w.r.t. the context-free case is coped with at compile time; for example, if there is one rule for nonterminal $A$:

$$A(x_1, x_2) \to x_3{:}B_1(x_1) \ B_2() \ B_3(x_3)/x_2$$

then the code for $[A](i, (l_1, r_1), (l_2, r_2))$ will be

    *result* := **empty**
    **for** $k_1 \in [B_1](i, (l_1, r_1))$
    **do**   $l_3 := i$
        $r_3 := k_1$
        **for** $k_2 \in [B_2](k_1)$
            **for** $k_3 \in [B_3](l_2, (l_3, r_3))$
                **if** $(k_3 == r_2)$
                      **add** $k_2$ **to** *result*
    **return** *result*

The extra effort remaining at parse time is in copying arguments and an occasional extra comparison (the **if** statement in the example), taking $m(1 + p)$ steps everytime the innermost **for** statement is reached, and the fact that not $\mathcal{O}(n)$, but $\mathcal{O}(n^{1+2p})$ argument-value pairs need to be memoized. Merging the results in a RHS sequence of $m$ items can be done in $\mathcal{O}(m(1 + p)n^{m-1})$ time. The result is a set of $\mathcal{O}(n)$ size. As there are at most $\mathcal{O}(|G|n^{1+2p})$ results to be computed, the overall time complexity of the algorithm is $\mathcal{O}(|G|m(1+p)n^{1+m+2p})$. □

■ **3.6 Remark** If all nonterminals in the grammar are nullary ($p = 0$), then the complexity result coincides with the values found for the context-free recursive descent algorithm (Leermakers, 1993). Nullary **LMG** includes the context-free case, but still allows movement local to a rule; the closure result 3.1 still holds for this class of grammars. As all we can do with binding and slashing local to a rule is intersection, the nullary **LMG**s must be precisely the closure of the context-free grammars under finite intersection.

These results can be extended to more efficient algorithms which can cope with left-recursive grammars such as memoizing *recursive ascent* (Leermakers, 1993). A very simple improvement is obtained by bilinearizing the grammar (which is possible if it is left binding), giving a worst case complexity of $\mathcal{O}(|G|(1+p)n^{3+2p})$.

## 4 Other Approaches to Separation of Movement

A natural question to ask is whether the **LMG** formalism (for the purpose of embedding in equational specification systems, or eliminating unification as a stage of sentence processing) really has an advantage over existing mildly context-sensitive approaches to movement. Other non-concatenative formalisms are head-wrapping grammars (**HG**) (Pollard, 1984), extraposition grammars (**XG**) (Pereira, 1981) and various exotic forms of tree adjoining grammar (Kroch and Joshi, 1986). For overviews see (Weir, 1988), (Vijay-Shanker et al., 1986) and (van Noord, 1993). The most applicable of these formalisms for our purposes seem to be **HG** and **XG**, as both of these show good results in modeling movement phenomena, and both are similar in appearance to context-free grammars; as in **LMG**, a context-free grammar has literally the same representation when expressed in **HG** or **XG**. Hence it is to be expected that incorporating these approaches into a system based on a context-free front-end will not require a radical change of perspective.

### 4.1 Head Grammars

A notion that plays an important role in various forms of Linguistic theory is that of a *head*. Although there is a great variation in the form and function of heads in different theories, in general we might say that the head of a constituent is the *key component* of that constituent. The *head grammar* formalism, introduced by Pollard in (Pollard, 1984) divides a constituent into three components: a *left context*, a terminal *head* and a *right context*. In a **HG** rewrite rule these parts of a constituent can be addressed separately when building a constituent from a number of subconstituents.

An accurate and elegant account of Dutch cross-serial dependencies using **HG** is sketched in (Pollard, 1984). However, we have not been able to construct head grammars that are able to model verb movement, cross-serial dependencies and topicalization at the same time. For every type of constituent, there is only *one* head, and hence only one element of the constituent that can be the subject to movement.[3]

### 4.2 Extraposition Grammars

Whereas head grammars provide for an account of verb fronting and cross-serial dependencies, Pereira, introducing *extraposition grammars* in (Pereira, 1981), is focused on displacement of noun phrases in English. Extraposition grammars are in appearance very similar to context-free grammars, but allow for larger patterns on the left hand side of PS rules. This makes it possible to allow a topicalized *NP* only if somewhere to its right there is an unfilled trace:

$$S \rightarrow \text{Topic } S$$
$$\text{Topic} \ldots XP \rightarrow NP$$

While **XG** allows for elegant accounts of cross-serial dependencies and topicalization, it seems again hard to simultaneously account for verb and noun movement, especially if the *bracketing constraint* introduced in (Pereira, 1981), which requires that **XG** derivation graphs have a planar representation, is not relaxed.[4]

Furthermore, the practical application of **XG** seems to be a problem. First, it is not obvious how we should interpret **XG** derivation graphs for further analysis. Second, as Pereira points out, it is nontrivial to make the connection between the **XG** formalism and standard (e.g. Earley-based) parsing strategies so as to obtain truly efficient implementations.

## 5 Conclusions

We have presented the **LMG** formalism, examples of its application, and a complexity result for a constrained subclass of the formalism. Example 2.9 shows that an **LMG** can give an elegant account of movement phenomena. The complexity result 3.5 is primarily intended to give an indication of how the recognition problem for **LMG** relates to that for arbitrary context free grammars. It should be noted that the result in this paper only applies to *non-combinatorial* **LMG**s, excluding for instance the grammar of example 2.9 as presented here.

There are other formalisms (**HG** and **XG**) which provide sensible accounts of the three movement phenomena sketched in section 1.3, but altogether do not seem to be able to model all phenomena at once. In (Groenink, 1995b) we give a more detailed analysis of what is and is not possible in these formalisms.

### Future Work

**1.** The present proof of polynomial complexity does not cover a very large class of literal movement grammars. It is to be expected that larger, Turing complete, classes will be formally intractable but behave reasonably in practice. It is worthwhile to look at possible practical implementations for larger classes of **LMG**s, and investigate the (theoretical and practical) performance of these systems on various representative grammars.

**2.** Efficient treatment of **LMG** strongly depends on the *left-binding* property of the grammars, which seems to restrict grammars to treatment of *leftward extraposition*. In reality, a smaller class of *rightward* movement phenomena will also need to be treated. It is shown in (Groenink, 1995b) that these can easily be circumvented in left-binding **LMG**, by introducing artificial, "parasitic" extraposition.

---

[3] However, a straightforward extension of head grammars defined in (Groenink, 1995a) which makes use of arbitrary tuples, rather than dividing constituents into three components, is (1) capable of representing the three target phenomena of Dutch all at once and (2) weakly equivalent to a (strongly limiting) restriction of literal movement grammars. Head grammars and their generalizations, being *linear context-free rewriting systems* (Weir, 1988), have been shown to have polynomial complexity.

[4] Theoretically simultaneous treatment of the three movement phenomena is not impossible in **XG** (a technique similar to *pit-stopping* in **GB** allows one to wrap extrapositions over natural bracketing islands), but grammars and derivations become very hard to understand.

## Acknowledgements

I would like to thank Jasper Kamperman, René Leermakers, Jan van Eijck and Eelco Visser for their enthousiasm, for carefully reading this paper, and for many general and technical comments that have contributed a great deal to its consistency and readability.